\documentclass[sigconf]{acmart}

\AtBeginDocument{%
  \providecommand\BibTeX{{%
    \normalfont B\kern-0.5em{\scshape i\kern-0.25em b}\kern-0.8em\TeX}}}

\copyrightyear{2021}
\acmYear{2021}
\setcopyright{acmcopyright}\acmConference[CIKM '21]{Proceedings of the 30th ACM
International Conference on Information and Knowledge Management}{November
1--5, 2021}{Virtual Event, QLD, Australia}
\acmBooktitle{Proceedings of the 30th ACM International Conference on Information
and Knowledge Management (CIKM '21), November 1--5, 2021, Virtual Event, QLD,
Australia}
\acmPrice{15.00}
\acmDOI{10.1145/3459637.3482492}
\acmISBN{978-1-4503-8446-9/21/11}

\usepackage{subfigure}
\usepackage{enumitem}
\usepackage{multirow}
\usepackage{algorithm}
\usepackage{algorithmicx}
\usepackage{bm}
\usepackage{algpseudocode}
\usepackage{balance}
\usepackage{amsmath}

\begin{document}

\title[HAIL]{What is Next when Sequential Prediction Meets \\Implicitly Hard Interaction?}
\fancyhead{}
\author{Kaixi Hu}
\affiliation{%
  \institution{Wuhan University of Technology}
  }
\email{issac_hkx@whut.edu.cn}

\author{Lin Li}
\authornote{Lin Li is the corresponding author.}
\affiliation{%
  \institution{Wuhan University of Technology}
  }
\email{cathylilin@whut.edu.cn}

\author{Qing Xie}
\affiliation{%
  \institution{Wuhan University of Technology}
  }
\email{felixxq@whut.edu.cn}

\author{Jianquan Liu}
\affiliation{%
  \institution{NEC Corporation}
  }
\email{jqliu@nec.com}

\author{Xiaohui Tao}
\affiliation{%
  \institution{University of Southern Queensland}
  }
\email{Xiaohui.Tao@usq.edu.au}

\renewcommand{\shortauthors}{Hu et al.}

\begin{abstract}
Hard interaction learning between source sequences and their next targets is challenging, which exists in a myriad of sequential prediction tasks. During the training process, most existing methods focus on explicitly hard interactions caused by wrong responses. However, a model might conduct correct responses by capturing a subset of learnable patterns, which results in implicitly hard interactions with some unlearned patterns. As such, its generalization performance is weakened. The problem gets more serious in sequential prediction due to the interference of substantial similar candidate targets.

To this end, we propose a \underline{H}ardness \underline{A}ware \underline{I}nteraction \underline{L}earning framework (HAIL) that mainly consists of two base sequential learning networks and mutual exclusivity distillation (MED). The base networks are initialized differently to learn distinctive view patterns, thus gaining different training experiences. The experiences in the form of the unlikelihood of correct responses are drawn from each other by MED, which provides mutual exclusivity knowledge to figure out implicitly hard interactions. Moreover, we deduce that the unlikelihood essentially introduces additional gradients to push the pattern learning of correct responses. Our framework can be easily extended to more peer base networks. Evaluation is conducted on four datasets covering cyber and physical spaces. The experimental results demonstrate that our framework outperforms several state-of-the-art methods in terms of top-k based metrics.
\end{abstract}




\maketitle

\section{Introduction}
\begin{figure}[t]
  \centering
  	\subfigure[Different types of interactions between source sequences and the next targets, and their divisions by jointly considering model responses and hardness.] {\label{figure1-1}
    	\includegraphics[width=0.9\columnwidth]{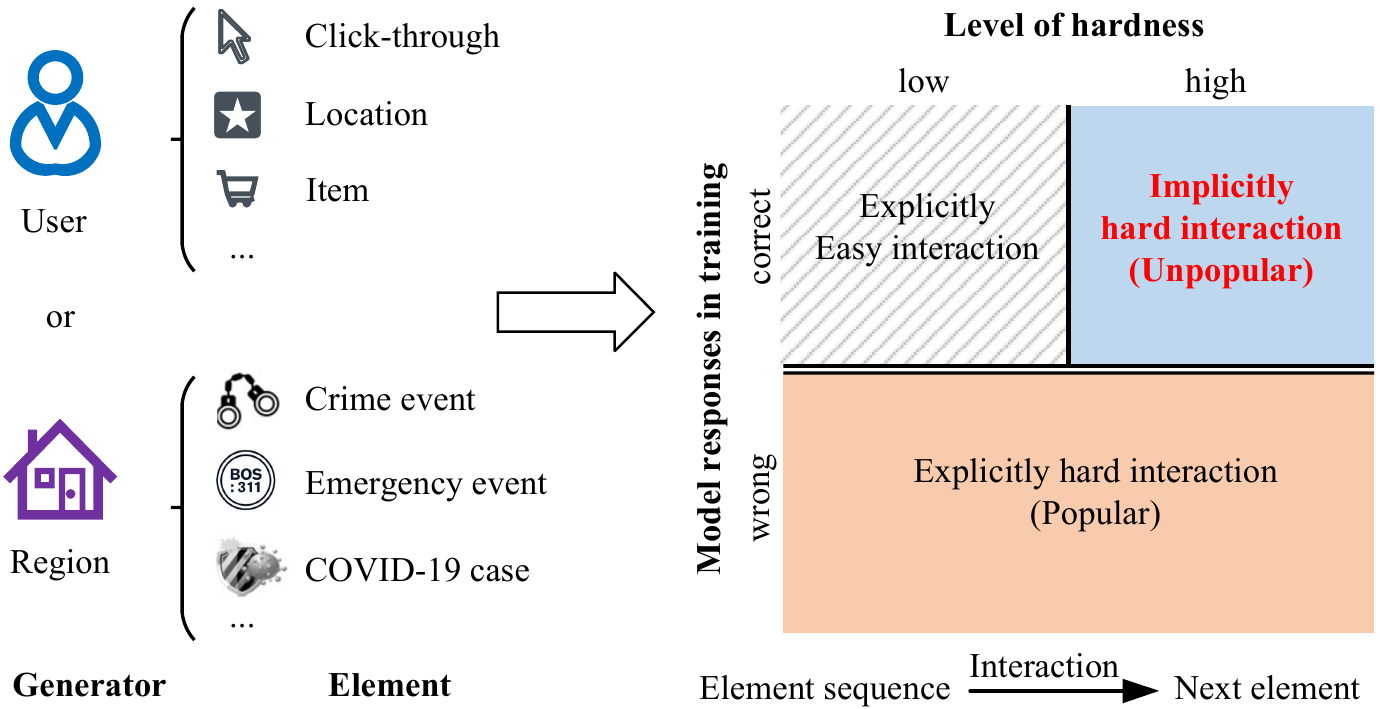}
    }
    \subfigure[Distinctive self-knowledge is generated from different initializations. The deeper the color, the higher likelihood of correct response.] {\label{figure1-2}
    	\includegraphics[width=0.95\columnwidth]{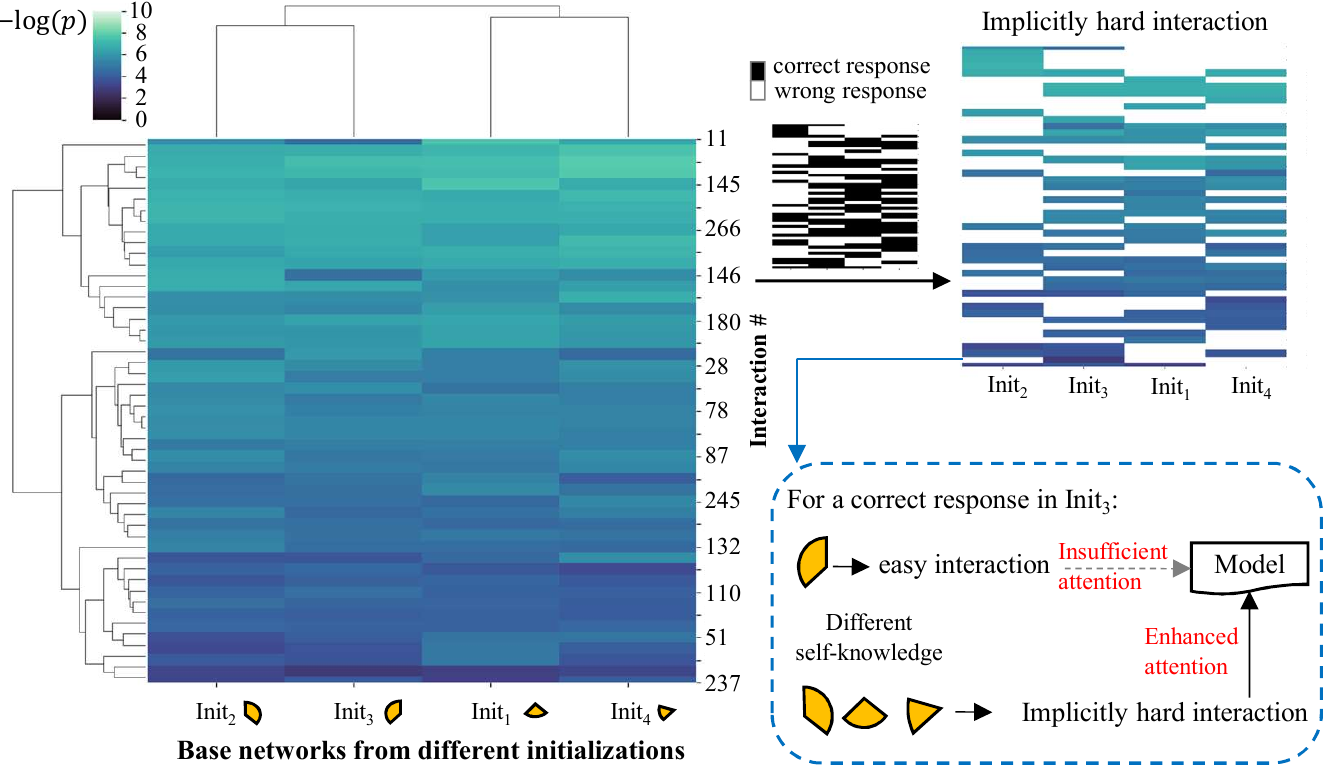}
    }
    \caption{An illustration of implicitly hard interactions caused by subsets of learnable patterns during training.}
  \label{figure1}
\vspace{-1.7 em}
\end{figure}

In modern society, various sequential prediction tasks can help humans making informed decisions such as recommendation \cite{DBLP:conf/wsdm/TangW18, DBLP:conf/cikm/SunLWPLOJ19}, trajectory prediction \cite{DBLP:conf/kdd/LianWG0C20}, click-through rate prediction \cite{DBLP:conf/kdd/PiBZZG19, DBLP:conf/dasfaa/ZengCZTMLZ20} and region-centered event prediction \cite{hu2021duronet}. For example, during the COVID-19 pandemic, precise prediction of infected cases has led to better allocation of healthcare resources~\cite{yan2020interpretable, swapnarekha2020role}. As shown in the left side of Figure \ref{figure1-1}, such tasks typically arrange a series of historical elements (e.g., items, events, locations or their counts) from a certain generator in chronological order, which constitutes a sequence. Existing studies have proved that there exist diverse kinds of interaction patterns between the element sequences and their corresponding next elements \cite{DBLP:conf/cikm/ZhouWZZWZWW20, DBLP:conf/kdd/LianWG0C20, DBLP:journals/is/LiCDWSX20, 9119847}. These patterns can give humans a hint about what the future element is like to some degree. In order to provide high-quality predictions, numerous researchers are trying to model and learn the latent interaction patterns comprehensively.

To capture the interaction patterns, various deep learning models are proposed. Most works notice that some interactions involved by humans present more complex characteristics such as irregular interaction \cite{DBLP:conf/iclr/ShuklaM19, DBLP:conf/wsdm/TangW18, DBLP:conf/cikm/SunLWPLOJ19}, dynamic dependency \cite{DBLP:conf/sigir/WangDH0C20, DBLP:conf/aaai/ZhouMFPBZZG19}, noise interference \cite{DBLP:conf/nips/LiJXZCWY19, hu2021duronet}, than others in nature. Therefore, the basic paradigm of their works is to design a well-adapted structure to better model sequential interactions. Recent studies indicate that an effective training strategy can also help to improve prediction performance by making use of interaction information \cite{DBLP:conf/cikm/ZhouWZZWZWW20, DBLP:conf/www/Yuan0JGXXX20, DBLP:conf/wsdm/WangF0NC21}. Despite the effectiveness of prior methods, most of them capture interaction patterns by revising wrong responses where ground-truth is not inferred correctly.

By jointly considering model responses in training and hardness of interactions, the mining of interactions can be divided into three types. As shown in the right part of Figure~\ref{figure1-1}, for the lower red region, the wrong responses often with relatively larger training loss are a kind of explicitly hard interactions that have caught the eye of researchers~\cite{DBLP:conf/cikm/SunLWPLOJ19, hu2021duronet, DBLP:conf/cikm/ZhouWZZWZWW20}. For the upper region, correct responses usually generate relatively lower loss \cite{DBLP:conf/wsdm/WangF0NC21}. However, among correct responses, there are kinds of interactions with some patterns unlearned by models. Low loss might make models no longer able to make specific intentional adjustments and prone to a subset of comprehensive patterns. With the limitation of training samples, how to learn such implicitly hard interactions is a burning problem to improve the generalization performance of sequential prediction models. More recently, a multi-view theory~\cite{allenzhu2020understanding} proposed by Microsoft Research indicates that individual models have limited capability of capturing multiple views of data. In addition, according to the biased assessment in some psychological mechanisms \cite{brown1991accuracy, karpen2018social}, one-sided perspective is easy to generate biased self-knowledge without communicating with others. Inspired by the above studies, multi-perspective experiences about pattern learning can be introduced to enhance the learning of implicitly hard interactions in sequential prediction. We think that some implicitly hard interactions can be identified from the inconsistent response results across different models.

To further investigate the implicitly hard interactions, we analyze the negative log-likelihood of 50 training samples in ML-1m (a movie ratings dataset) by a hierarchically-clustered heatmap. As shown in Figure~\ref{figure1-2}, several observations are listed as follows:
\begin{itemize}[leftmargin=*, topsep = 0 pt]
\item From the left side, the likelihood of predicting positive samples varies with initializations. And, both the interactions and initializations present cluster effects by observing the dendrogram on the left and top of heatmap, respectively. It denotes that models with different initializations can generate distinctive self-knowledge.
\item After masking the wrong responses with white blocks, inconsistent training results accounted for 15.42\% of total samples can be observed from the right side. It is noted that some interactions are correctly predicted with high likelihood (deep color) by only a model while other models present wrong responses.
\end{itemize}
Overall, the observations are in line with the limitation of self-knowledge \cite{wilson2004self, brown1991accuracy, karpen2018social, allenzhu2020understanding}. And, inspired by them, multiple perspectives of interactions can be generated by distinctive initializations.

To this end, we propose a novel Hardness Aware Interaction Learning framework (HAIL). Our solution aims at exchanging mutual exclusivity knowledge, which aggregates training experiences from other perspectives for learning implicitly hard interactions. \emph{First}, two base networks are developed from different initializations to generate distinctive information about implicitly hard interactions.  \emph{Second}, we propose mutual exclusivity distillation (MED) that subtly transfers the unlikelihood of correct responses for different interactions. We further infer that such mutual exclusivity knowledge in the form of unlikelihood can adjust the gradients of models, which can enhance the learning of implicitly hard interactions. This learning paradigm is conducive to improving generalization performance of models by enriching view patterns and can be easily extended to more peers.

Our main contributions are summarized as follows:
\begin{itemize}[leftmargin=*]
\item We highlight the impact of implicitly hard interactions and identify their inconsistent characteristics across different perspectives. To our knowledge, it is the first time that implicitly hard interactions are mentioned in sequential prediction tasks.
\item A general learning framework HAIL is developed for sequential prediction to enhance the learning of implicitly hard interactions. In particular, MED is proposed to derive mutual exclusivity knowledge, which breaks the conventional manner of mimic learning in knowledge distillation. We further infer that MED essentially introduces additional gradients to push pattern learning of implicitly hard interactions.
\item With extensive experiments on two benchmark recommendation datasets from cyber space and two event datasets from physical space, the proposed framework HAIL outperforms existing state-of-the-art methods in several typical applications in terms of top-k based metrics.
\end{itemize}

\section{Related Work}
In this section, we review some sequential prediction works with respect to hard interactions, and then investigate the advance of knowledge distillation.

\subsection{Sequential Prediction}
Sequential prediction is a common technique that is widely used in various domains such as sequential recommendation \cite{DBLP:conf/wsdm/TangW18, DBLP:conf/cikm/SunLWPLOJ19}, location prediction \cite{DBLP:conf/kdd/LianWG0C20}, click-through rate prediction~\cite{DBLP:conf/kdd/PiBZZG19, DBLP:conf/dasfaa/ZengCZTMLZ20}.

\textbf{\emph{Incorporating More Information.}} Early matrix factorization based methods \cite{DBLP:conf/ijcai/ChengYLK13, DBLP:conf/recsys/JuanZCL16, DBLP:conf/www/HeLZNHC17} are difficult to capture hard interactions. With the emerging of deep learning, substantial neural networks are proposed to learn hard interactions by incorporating more information, such as RNN \cite{DBLP:conf/sigir/YuLWWT16, DBLP:conf/icdm/LiSZ18, DBLP:conf/ijcai/ChengYLK13} and convolution \cite{DBLP:conf/wsdm/TangW18, BaiTCN2018}. Recently, since self-attention~\cite{DBLP:conf/nips/VaswaniSPUJGKP17} shows promising performance, more advanced self-attention based methods are proposed to introduce related information in terms of different applications. These methods can be divided into two aspects: 1) For the tasks in cyber space, recommendation is one of the hotspot research areas \cite{DBLP:conf/sigir/WangDH0C20, DBLP:conf/cikm/ZhouWZZWZWW20, DBLP:conf/cikm/SunLWPLOJ19}; 2) For the tasks in physical space,  there are convolution kernels \cite{DBLP:conf/nips/LiJXZCWY19}, sparse mechanism based Informer \cite{informer2021}, geography-aware based GeoSAN \cite{DBLP:conf/kdd/LianWG0C20} and adjacent context based DuroNet \cite{hu2021duronet}.
%

\textbf{\emph{Learning Strategy.}} More recently, some studies indicate that effective training strategies can also enhance the learning of hard interactions and they are less affected by specific applications. MIM can well capture intrinsic data correlation to avoid overemphasizing the final performance \cite{DBLP:conf/cikm/ZhouWZZWZWW20}. On the premise of sufficient data, the learning of partial hard interactions can be amplified by removing noisy interactions \cite{DBLP:conf/wsdm/WangF0NC21}.

The above methods focus on learning hard interactions under self-knowledge from a single perspective, which pays less attention to implicitly hard interactions. Different from them, our work tries to capture implicitly hard interactions no matter which specific cyber or physical spaces they are in by deriving training experience from others. To this end, we propose a novel hardness aware interaction learning framework that mainly consists of two base networks and a MED strategy.

\subsection{Knowledge Distillation}
Given a training model, this paper focuses on how to draw experiences about learning implicitly hard interactions from other models. Knowledge distillation is an effective means to transfer knowledge between models, which is particularly suitable in our scenario.

\textbf{\emph{Model Compression.}} Ideas underpinning distillation can date back to model compression \cite{DBLP:conf/kdd/BucilaCN06}. The current and most well-known neural distillation is proposed by Hinton et al. \cite{DBLP:journals/corr/HintonVD15} where small student models can derive more information from the softened output of cumbersome teacher models than the ground-truth. Subsequently, a long line of papers about distillation and compression quickly emerges. In sequential prediction, some progress has been made in the \emph{distillation object, architectures and procedures}, respectively. Specifically, some features from the teacher's hidden layer are also distilled to guide the learning of student models \cite{DBLP:journals/corr/RomeroBKCGB14}. Self-distillation allows the teacher and students lying in a same model architecture \cite{DBLP:conf/acl/LiuZWZDJ20}. A two-step distillation is proposed for the pre-training and fine-tuning stage, respectively \cite{DBLP:conf/emnlp/JiaoYSJCL0L20}.

\textbf{\emph{Non-compression Task.}} Recently, distillation is proved to be feasible in other tasks that are not for model compression. Two typical works break the learning pattern from teacher models to improve performance of image classification. Born-again neural network obtains improvement from the prior model by teaching selves \cite{DBLP:conf/icml/FurlanelloLTIA18}. Deep mutual learning collects knowledge from student cohort \cite{DBLP:conf/cvpr/ZhangXHL18}. Moreover, multilingual translation can be integrated into a unified model by distilling different language models \cite{DBLP:conf/iclr/TanRHQZL19}.

In this work, we break the mimic learning in convention where student models try to reproduce the knowledge from their teacher models. We argue that imitation is not much appropriate to further improve performance, since the teacher models preferentially transfer selective knowledge that they are in high confidence. To this end, MED is proposed to employ mutual exclusivity knowledge that is also a kind of learning experience. In this manner, a model can acquire hints from the unlikelihood of others' correct responses to notice implicitly hard interactions.

\section{Framework}
In sequential prediction, the principal entities are generators (e.g., users, regions) and elements (e.g., items, locations, events). The generator can generate a series of elements in chronological order. Given a set of generators $\mathcal{G}=\{g_1, g_2, ...,g_{\vert\mathcal{G}\vert}\}$ and a set of elements $\mathcal{E}=\{e_1, e_2, ...,e_{\vert\mathcal{E}\vert}\}$, the interaction sequence in chronological order for generator $g \in\mathcal{G}$ can be denoted as $\mathcal{X}_g=\{e^{(g)}_1, ..., e^{(g)}_t, ..., e^{(g)}_{n_g}\}$, where $e^{(g)}_t \in \mathcal{E}$ is the element that $g$ has interacted with at time step $t$ and $n_g$ is the length of interaction sequence for generator $g$.

\textbf{Sequential Prediction for Next.} Based on the above notations, the task of sequential prediction can be formally defined as follows: given the historical interaction sequence $\mathcal{X}_g$, the objective is to learn a prediction model $\mathcal{M}: \mathcal{X}_g \rightarrow \mathcal{P}_g$ where $\mathcal{P}_g$ is the likelihood distribution over all elements that generator $g$ possibly interacts with at time step $n_g+1$. The next element $\hat{\mathcal{Y}}$ can be inferred by sorting the likelihood in descending order.

\subsection{Overview}
\begin{figure*}[t]
  \centering
  \includegraphics[width=0.94\linewidth]{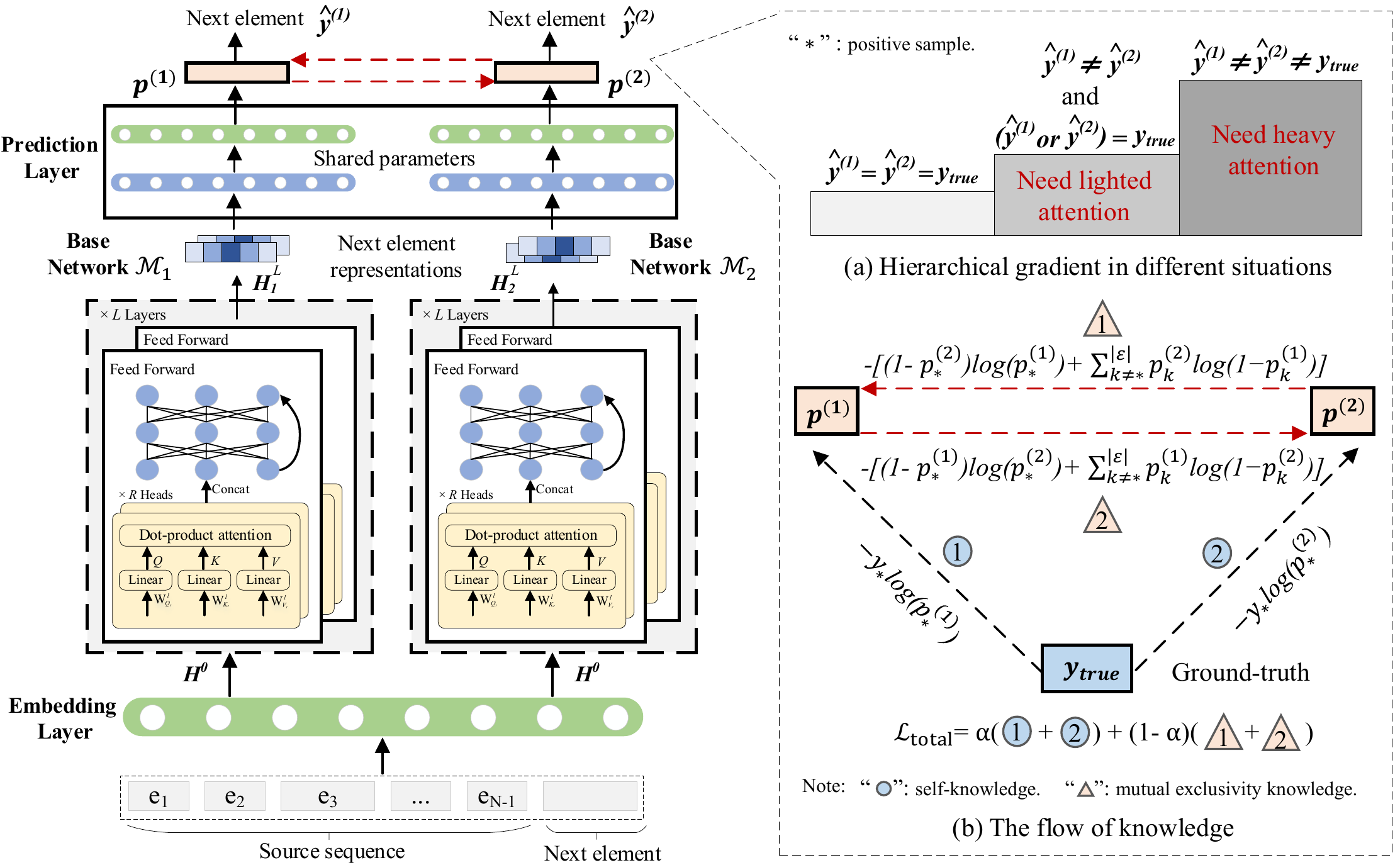}
	  \caption{The framework of the proposed HAIL. Implicitly hard interactions are identified by the inconsistent results of two base networks as shown in Subfigure (a). The base networks exchange mutual exclusivity knowledge in the form of unlikelihood to enhance the learning of implicitly hard interactions as shown in Subfigure (b).}
  \label {framework}
\end{figure*}

The framework of HAIL is presented in Figure \ref{framework}.
HAIL consists of a shared interaction embedding layer, two base networks (peer for each other) and a shared prediction layer. In the training stage, the experience of learning implicitly hard interactions can be distilled from the peer model. In prediction, either of the base networks can be removed.

The basic idea of our work is to employ the mutual exclusivity knowledge of peer's correct responses to enhance the learning of implicitly hard interactions. To this end, two base networks from different initializations are designed to generate distinctive self-knowledge. Then, MED is proposed to make base networks close to each other and exchange the unlikelihood of correct responses. Finally, each model collects the self-knowledge and the mutual exclusivity knowledge from its peer to adjust the learning weights for implicitly hard interactions.

In the following, we first introduce the components of the proposed framework HAIL. And then, we elaborate on the hardness aware learning and the proposed MED.

\subsection{Interaction Embedding Layer}
As shown in the lower-left region of Figure \ref{framework}, a sequence of elements (a.k.a. interaction) is first embedded into fixed-length vectors by looking up a shared embedding table $\bm{U} \in \mathbb{R}^{{\vert\mathcal{E}\vert} \times d}$ where ${\vert\mathcal{E}\vert}$ is the number of elements and $d$ is the length of vectors. Here, we train the table from scratch without introducing any pre-training parameters. To make use of positional information, we also employ a positional table $\bm{S} \in \mathbb{R}^{N \times d}$ to generate a fixed-length position vector where $N$ is the maximum length of input sequence. Formally, the output of embedding layer can be derived by suming the element embedding and the positional embedding as $\bm{h^0_i} = \bm{x_i}\bm{U}+\bm{s_i}$, where $\bm{x_i}$, $\bm{s_i}$ and $\bm{h^0_i}$ are one-hot input, position and embedding vector of the $i$th element, respectively. The embedding matrix $\bm{H^0}=[\bm{h^0_1},...,\bm{h^0_i},...\bm{h^0_N}]$ will be fed into the following components.

\subsection{Base Network}
As shown in the middle-left region of Figure \ref{framework}, to introduce additional knowledge, two base networks initialized differently are employed to independently model interactions. We attempt to capture multi-perspective interactions by employing dark knowledge~\cite{DBLP:conf/cvpr/ZhangXHL18}, since models learn from different starting points can derive mutable probability distributions. Note that the architecture of base networks can be set flexibility according to specific applications. Without loss of generality, we do not distinguish network $\mathcal{M}_1$ and network $\mathcal{M}_2$ in the following description for simplicity.

\subsubsection{Multi-head Self-attention Layer}
In most sequential prediction tasks, interactions hide in a relatively long-time span. The conventional RNNs are easy to meet the vanishing gradient problem~\cite{DBLP:conf/nips/SutskeverVL14}. Hence, we adopt the self-attention that captures interactions between elements without regard to distance.

In particular, the input of the $l$th layer $\bm{H^{l-1}}$ is transformed into $R$ subspaces simultaneously to derive $R$ attention heads. Then, the heads are concatenated and transformed again to output the representations, after a residual connection. The process can be defined as follows:
\begin{equation}
\begin{split}
\bm{A_0^l}&=Concat(\bm{head_1},...,\bm{head_r},...,\bm{head_R})\bm{W_O^l}+\bm{H^{l-1}},\\
&\bm{head_r}=Attention(\bm{H^{l-1}}\bm{W_{Q_r}^{l}},\bm{H^{l-1}}\bm{W_{K_r}^{l}},\bm{H^{l-1}}\bm{W_{V_r}^{l}}),
\end{split}
\end{equation}
where $\bm{W_{Q_r}^{l}},\bm{W_{K_r}^{l}},\bm{W_{V_r}^{l}} \in \mathbb{R}^{d \times d/R}$ are three projection matrices of the $r$th subspace. $\bm{W^l_O}\in\mathbb{R}^{d \times d}$ is the output projection. $R$ is the number of heads. $\bm{A^l_0}$ is the final representation after residual connection. The attention function is a scaled dot-product computation that can be calculated as
\begin{equation}
Attention(\bm{Q},\bm{K},\bm{V})=softmax(\frac{\bm{Q}\bm{K^T}}{\sqrt{d/R}})\bm{V}.
\end{equation}

\subsubsection{Feed Forward layer}
To introduce more nonlinearity, a two-layer feed forward network is applied on each element representations as follows:
\begin{equation}
\begin{split}
  \bm{A^l_1}&=\tau(\bm{A^l_0}\bm{W^l_1}+\bm{b^l_1}),\\
  \bm{A^l_2}&=\bm{A^l_1}\bm{W^l_2}+\bm{b^l_2},
\end{split}
\end{equation}
where $\bm{W^l_1}\in \mathbb{R}^{d \times d_h}$ and $\bm{W^l_2}\in \mathbb{R}^{d_h \times d}$ are weight matrixes. $\bm{b^l_1}\in \mathbb{R}^{d_h}$ and $\bm{b^l_2}\in \mathbb{R}^{d}$ are the biases. $d_h$ is the dimension of the intermediate layer. $\tau(\cdot)$ is the activation function (GELU in our experiment). Finally, the output of the $l$th encoder can be derived after a residual connection as $\bm{H^{l}}=\bm{A^l_2}+\bm{A^l_0}$.

\subsection{Prediction Layer for Next}
As shown in the upper-left region of Figure \ref{framework}, the final representations of masked elements are fed into a prediction layer after iteratively computing of the latent patterns. A shared feed forward and the shared embedding table are employed to decode them. The likelihood distribution $\bm{p^{(j)}}$ of the predicted next element for the $j$th base network can be derived as follows:
\begin{equation}
\begin{split}
  \bm{z}^{(j)}&=\tau(\bm{h^{(j,L)}}\bm{W_P}+\bm{b_P})\bm{U^T}+\bm{b_U},\\
  \bm{p^{(j)}}&=\Psi(\bm{z^{(j)}}),
\end{split}
\label{pj}
\end{equation}
where $\bm{h^{(j,L)}}$ denotes the final representation of the $L$th layer and $L$ is the number of layers, $\bm{W_P}\in \mathbb{R}^{d\times d}$ is the weight matrix, $\bm{b_P}\in\mathbb{R}^{d}$ and $\bm{b_U}\in\mathbb{R}^{\vert\mathcal{E}\vert}$ are the biases.  $\bm{z^{(j)}}=[z^{(j)}_1,...,z^{(j)}_{\vert\mathcal{E}\vert}]$ is the output of logits. $\Psi(\cdot)$ is the score function (softmax in our experiment).

\subsection{Hardness Aware Learning}
As shown in right part of Figure \ref{framework}, for each base network, the ground-truth and the likelihood distribution of the output from its peer are vital sources of knowledge. To make use of them, self-knowledge independent learning and mutual exclusivity knowledge distillation are designed, respectively.

\subsubsection{Self-knowledge Independent Learning}
To obtain a decent baseline and avoid model drifting arbitrarily, both base networks are designed to learn from ground-truth independently. In this way, they can acquire distinctive self-knowledge from different initial learning points. By following most existing methods~\cite{DBLP:conf/naacl/DevlinCLT19, DBLP:conf/cikm/ZhouWZZWZWW20, DBLP:conf/cikm/SunLWPLOJ19}, self-supervised learning is adopted in our work. In particular, for any element sequence, we randomly mask a proportion of elements with special tokens "[Mask]". This process can be repeated multiple times to generate multiple masked sequences. It is worth noting that more training sequences with final elements masked are appended to avoid fine-tuning in prediction \cite{DBLP:conf/cikm/SunLWPLOJ19}.

As the blue circle shown in the lower-right region of Figure \ref{framework}, the cross-entropy loss is adopted to converge the proposed model. For each input sequence, the self-knowledge based loss of the $j$th base model can be defined as:
\begin{equation}
  \mathcal{L}_{SK}^{(j)} = -\frac{1}{{\vert\mathcal{I}\vert}} \sum_{i \in \mathcal{I}} y_{i,*}log(p^{(j)}_{i,*}),
\label{ik}
\end{equation}
where $\mathcal{I}$ is the index set of masked elements in the input sequence. For the $i$th masked element, $\bm{y_{i}}=[y_{i,1}, ..., y_{i,*},...,y_{i,{\vert\mathcal{E}\vert}}]$ is the corresponding ground-truth label. $\bm{p^{(j)}_{i}}=[p^{(j)}_{i,1}, ..., p^{(j)}_{i,*},...,p^{(j)}_{i,{\vert\mathcal{E}\vert}}]$ is the likelihood distribution of all elements. $*$ denotes the positive sample.

\subsubsection{Mutual Exclusivity Knowledge Distillation.}
\label{HIL}
When training a model, learning experiences from its peer can be introduced by knowledge distillation. For conventional mimic learning \cite{DBLP:journals/corr/HintonVD15, DBLP:conf/cvpr/ZhangXHL18}, the student model preferentially obtains the experiences with high likelihood. However, for sequential prediction tasks, the likelihood of hard interactions is generally not on a high level. Hence, such knowledge from mimic learning is not suitable for our tasks. To address the problem, we employ another learning experience that is mutual exclusivity knowledge of correct responses to enhance the learning of implicitly hard interactions.

As the red triangle shown in the right bottom part of Figure~\ref{framework}, the mutual exclusivity knowledge based loss is derived by employing the posterior unlikelihood of correct responses from the peer network. For each training sequence, the loss of the $j$th base network can be defined as:
\begin{equation}
  \mathcal{L}_{MEK_{pos}}^{(j)} = -\frac{1}{{\vert\mathcal{I}\vert}} \sum_{i \in \mathcal{I}} (1-p^{(\neq j)}_{i,*})log(p^{(j)}_{i,*}),
\label{soft_pos}
\end{equation}
where $p^{(\neq j)}_{i,*}$ denotes the likelihood of positive label from the peer network. Note that $p^{(\neq j)}_{i,*}$ is the distillation target in conventional mimic learning \cite{DBLP:journals/corr/HintonVD15}. Differently, in our work, $(1-p^{(\neq j)}_{i,*})$ represents the mutual exclusivity knowledge that is distilled in the learning of positive samples.

Most sequential prediction tasks require predicting the next element from tens of thousands of candidates. It is intuitive that the highly similar elements might present more serious interference for the target. To this end, the mutual exclusivity knowledge is also introduced in the learning of negative samples. The sum of their loss is defined as:
\begin{equation}
  \mathcal{L}_{MEK_{neg}}^{(j)} = -\frac{1}{{\vert\mathcal{I}\vert}} \sum_{i \in \mathcal{I}} \sum_{k\neq *}^{\vert\mathcal{E}\vert} p^{(\neq j)}_{i,k}log(1-p^{(j)}_{i,k}),
	\label{soft_neg}
\end{equation}
where $p^{(j)}_{i,k}$ is the likelihood of the $k$th element in the likelihood distribution $\bm{p^{(j)}_{i}}$ and $p^{(\neq j)}_{i,k}$ is the likelihood of the $k$th element from the peer model. $*$ denotes the positive sample.

\subsubsection{Denoising} The interference of noisy interactions which do not reflect true preference \cite{DBLP:conf/wsdm/WangF0NC21}, is an inevitable problem when enhancing the learning of implicitly hard interactions. It is harmful to make model fit them, which may hurt the generalization \cite{DBLP:conf/icml/BengioLCW09, DBLP:conf/icml/JiangZLLF18}. Some works point out that the number of noisy interactions is less and their losses are larger \cite{DBLP:conf/aaai/Li0W19, DBLP:conf/wsdm/WangF0NC21}. As such, the shared interactions with large losses in the self-knowledge independent learning are truncated in the failure experience mutual learning. Formally, for each positive or negative sample, the truncated hard interaction loss is defined as follows:

\begin{equation}
  \bar{\mathcal{L}}_{MEK_{\{pos,neg\}}}^{(j)} = \left \{
\begin{array}{ll}
    0,  & rank(\mathcal{L}_{SK}^{(\forall j)})\textless \beta \cdot S_L\\
    \mathcal{L}_{MEK_{\{pos,neg\}}}^{(j)},  & otherwise,
\end{array}
\right.
\label{denoise}
\end{equation}
where $\beta$ is the proportion of truncated noisy interactions and $S_L$ is the size of interactions. rank($\cdot$) denotes the rank of the loss in all interactions in descending order. In Equation (\ref{denoise}), the noisy interactions are also identified by jointly conditioning on the losses from both networks.

\subsubsection{Loss Balance} To balance the knowledge between the ground-truth and the peer model, a balance factor $\alpha$ is introduced to derive the total loss of the $j$th base model as follows:
\begin{equation}
  \mathcal{L}_{total}^{(j)} = \alpha\mathcal{L}_{SK}^{(j)} + (1-\alpha)(\bar{\mathcal{L}}_{MEK_{pos}}^{(j)}+\bar{\mathcal{L}}_{MEK_{neg}}^{(j)}).
\label{total}
\end{equation}
Eventually, the final loss function that is adopted to converge the proposed model is defined as follows:
\begin{equation}
  \mathcal{L}_{total} = \mathcal{L}_{total}^1 + \mathcal{L}_{total}^2.
\end{equation}

\subsection{Extension to More Peers}
Notwithstanding the promising exclusivity knowledge transfer between two base networks, the proposed MED can be naturally extended to more peers with different parameters or structures. More base networks are expected to introduce such knowledge from diverse perspectives to conduct learning of implicitly hard interactions. And in the follow-up deployment, the redundant networks can be removed for reducing the computation.

Given $T$ ($T\ge 2$) base networks, the mutual exclusivity knowledge distillation can be extended as Equations (\ref{exmekpos}) and (\ref{exmekneg}). For each network, it can obtain $T-1$ hints that are introduced from the other peers. Such that, Equations~(\ref{soft_pos}) and (\ref{soft_neg}) can be regarded as a special situation with $T=2$. Here, a rescaling factor $\frac{1}{T-1}$ is introduced to ensure a balanced value of loss.

\begin{equation}
  \tilde{\mathcal{L}}_{MEK_{pos}}^{(j)} = -\frac{1}{T-1} \sum^T_{c \neq j}  (1-p^{(c)}_{*})log(p^{(j)}_{*}).
  \label{exmekpos}
\end{equation}
\begin{equation}
  \tilde{\mathcal{L}}_{MEK_{neg}}^{(j)} = -\frac{1}{T-1}  \sum^T_{c \neq j}  \sum_{k\neq *}^{\vert\mathcal{E}\vert} p^{(c)}_{k}log(1-p^{(j)}_{k}).
  \label{exmekneg}
\end{equation}

%
%

\section{Discussion}
The key contribution of our work is to enhance the learning of implicitly hard interactions by employing training experience. To obtain more insight about it, we explore how MED works. Without loss of generality, the discussion focuses on the model $\mathcal{M}_1$. We aim to answer the following questions:

\noindent\textbf{Question 1:} \emph{Why does MED work? What does mutual exclusivity knowledge distillation bring?}

During the exchange of training experience, the parameters are usually updated along the direction of negative gradient. Without loss of generality, for each positive sample in the base model $\mathcal{M}_1$, the gradient of cross-entroy \cite{DBLP:books/lib/Murphy12} in Equation (\ref{soft_pos}) with respect to base model's logits $z^{(1)}_*$ in Equation (\ref{pj}) can be derived as:
\begin{small}
\begin{equation}
\begin{split}
   &\frac{\partial {\mathcal{L}_{MEK_{pos}}^{(1)}}}{\partial {z^{(1)}_*}}=\frac{\partial {\mathcal{L}_{MEK_{pos}}^{(1)}}}{\partial {p^{(1)}_*}} \frac{\partial {p^{(1)}_*}}{\partial {z^{(1)}_*}}= -\frac{1-p_*^{(2)}}{p_*^{(1)}}\cdot p_*^{(1)}(1-p_*^{(1)})        \\
	&=\underbrace{ (1-p^{(2)}_{*})(p^{(1)}_{*}-y_*)}_{\textcircled{1} }=\underbrace{p^{(1)}_{*}-y_*}_{self-knowledge}+\underbrace{p^{(2)}_{*}(y_*-p^{(1)}_{*})}_{\textcircled{2}},
\end{split}
\label{gradient}
\end{equation}
\end{small}
where $y_*=1$ is the label of positive samples. $p^{(1)}_{*}$ and $p^{(2)}_{*}$ are the likelihood of the next element.

As shown in Equation (\ref{gradient}), the final gradient is rewritten as term~$\textcircled{1}$. In this form, $(1-p^{(2)}_{*})$ can be interpreted as an importance weight of the original ground-truth label $y_*$. When base model $\mathcal{M}_2$ makes a serious mistake (i.e., $p^{(2)}_{*}\approx 0$), Equation (\ref{gradient}) is approximate to the gradient generated by the self-knowledge based loss in Equation (\ref{ik}). In this situation, base model $\mathcal{M}_2$ gives base model $\mathcal{M}_1$ a hint that the current interaction is hard to learn. And, base model $\mathcal{M}_1$ will enhance the learning of the interaction with a gradient affected by another perspective. While base model $\mathcal{M}_2$ with a higher likelihood, the gradient of the interaction will be rescaled and generate less contribution. Along this line, MED can take similar effect on negative samples.

Compared with self-knowledge independent learning, mutual exclusivity knowledge distillation contains an additional term $\textcircled{2}$ in Equation (\ref{gradient}). The gradient of implicitly hard interaction is derived by jointly conditioning on the likelihood from distinctive initializations. More views of data are introduced to enhance the learning of implicitly hard interactions. As shown in the upper-right region of Figure~\ref{framework}, the combination of prediction results and ground-truth can be divided into different levels. In particular, if base models generate inconsistent results, a trade-off gradient can be derived. Therefore, mutual exclusivity knowledge distillation conduct a hierarchical learning of different interactions.

\noindent\textbf{Question 2:} \emph{What is the difference between conventional mimic learning (learn from $p^{(2)}_*$) and MED (learn from $1-p^{(2)}_*$)?}

It is straightforward that the proposed MED can dynamically rescale the importance weights of interactions by their wrong responses, which aims to distinguish the next target from substantial candidates. The conventional mimic learning carries information from teacher to students, which emphasises similar knowledge among similar elements. In this case, the weight~$(1-p^{(2)}_*)$ in Equation (\ref{gradient}) is replaced by $p^{(2)}_*$, which means the model prefers to easy interactions. Therefore, for substantial elements in sequential prediction, MED can effectively reduce the interference, which is more suitable in our scenario.

\section{Experiments}
In this section, experiments are conducted on different datasets to validate the effectiveness of our HAIL. In particular, we aim to answer the following research questions:
\begin{itemize}[leftmargin=*]
\item {\bfseries RQ1:} How does our HAIL perform compared with the state-of-the-art sequential prediction methods?
\item {\bfseries RQ2:} How is the performance of HAIL variants with different combinations of terms in loss function (Equation (\ref{total}))?
\item {\bfseries RQ3:} What is the effect of the truncation proportion $\beta$ in denoising (Equation (\ref{denoise})) and the balance factor $\alpha$ in loss function~(Equation~(\ref{total}))?
\end{itemize}

\subsection{Experimental Setup}
\subsubsection{Dataset}
Experiments are conducted on two benchmark recommendation datasets from cyber space and two crime datasets from physical space. These datasets that are involved by humans contain more implicitly hard interactions. As shown in Table \ref{dataset}, the size of different datasets varies with the domains.
\begin{itemize}[leftmargin=*, topsep = 0 pt]
\item{\textbf{ML-1m}\footnote{https://grouplens.org/datasets/movielens/1m/}:} This is a movie ratings dataset created in February, 2003. As one of the stable benchmark datasets, most recommendation algorithms are evaluated on it.
\item{\textbf{Toys}:} Toys is a subcategory in Amazon review dataset. We obtained this dataset from \cite{DBLP:conf/cikm/ZhouWZZWZWW20}.
\item{\textbf{CHI-18}\footnote{https://data.cityofchicago.org/Public-Safety/Crimes-2001-to-present/ijzp-q8t2}:} This is a public crime dataset updated by Chicago Police Department in 2018. To achieve a more fine-grained prediction, an event is described by a simple event model (SEM) \cite{DBLP:journals/ws/HageMSHS11}. In particular, SEM can model basic events in various domains without domain-specific vocabularies. In this work, the fields about geographical information are extracted to describe generators. Then, time slots are divided every 3 hours, which is smaller than the meaningful interval 6 hours \cite{nitsure2020unlocking}. The time slot and crime type are together used to describe elements.
\item{\textbf{NYC-16}\footnote{https://data.cityofnewyork.us/Public-Safety/NYPD-Complaint-Data-Current-Year-To-Date-/5uac-w243}:} This is a crime dataset provided by New York City Police Department in 2016. Similar to Chicago, the precincts and premises are used to describe generators. The time slots and classification codes are leveraged to describe elements.
\end{itemize}

\begin{table}
  \caption{Statistics of experimented datasets}
	\centering
\resizebox{0.9\linewidth}{!}{
  \begin{tabular}{ccccccc}
    \toprule
    \multirow{2}{*}{Dataset}&		\multirow{2}{*}{Generator \#}&		\multirow{2}{*}{Elements \#}&		\multicolumn{4}{c}{Sequence Length}\\
\cline{4-7}
		&			&		&		Max.&	Min.&		Avg.&			Std.			\\
	\midrule
    ML-1m& 		6,040&	3,416&		2,275&	16&		163.50&			192.53				\\
    Toys&		19,412&	11,924&		548&	3&		6.63&			8.50				\\
	\midrule
    CHI-18& 	2,692&	246&		3,525&	3&		96.18&			253.36			\\
    NYC-16& 	3,229& 	440&		4,496&	3&		144.76&			429.78			\\
  \bottomrule
\end{tabular}
\label {dataset}
}
\vspace{-1 em}
\end{table}

For all datasets, the elements are grouped by generators and sorted in chronological order for each generator. The inactive generators with fewer than five elements are removed to ensure the quality of prediction. Moreover, the last element in each sequence is taken as the test data and the element before the last element as the validation set. The remaining elements are used for training. The maximum length of sequence is set as 200. To ensure the sequence within the maximum length, longer sequences will be sliced into multiple sub-sequences from right to left.

\subsubsection{Metrics}
Following the common assessment means \cite{DBLP:conf/sigir/WangDH0C20, DBLP:conf/cikm/ZhouWZZWZWW20}, the performance of prediction can be assessed by top-$k$ Hit Ratio (HR$@k$), top-k Normalized Discounted Cumulative Gain (NDCG$@k$), and Mean Reciprocal Rank (MRR). In this paper, the cutoff $k$ is set as \{1,5,10\}. Note that HR@1 is equal to NDCG@1 that is a harsh metric for performance evaluation. To achieve an efficient computation in a large candidate set, 99 negative elements are randomly selected to rank with the target element. For all metrics, the higher the value, the better the~performance.

\subsubsection{Settings}
We implemented the proposed HAIL in Python with TensorFlow and conducted experiments on a commodity machine equipped with a 12GB TITAN Xp GPU. We train the model by using Adam with NOAM decay \cite{DBLP:conf/nips/VaswaniSPUJGKP17}. The batch size is set as 256. In base network, we set the layer number $L$ as 2, the head number $R$ as 2, the hidden size $d$ as 64, the intermediate size as 256. For learning hyper-parameters, $\beta$ is tuned in \{0,0.01,0.02,0.03\}, $\alpha$ is searched in \{0.1,0.2,...,0.8,0.9\}. The source code is available at GitHub\footnote{https://github.com/hukx-issac/HAIL}.

\subsection{Baselines}
To validate the effectiveness and generalization of our HAIL, we conduct a comparison with eight baselines from different domains. They are elaborated as follows:
\begin{itemize}[leftmargin=*, topsep = 0 pt]
\item {\bfseries POP}: A non-sequential baseline that simply regards the frequency of interactions as the probability of the next element.
\end{itemize}
\noindent{\bfseries Recommendation Methods:}
\begin{itemize}[leftmargin=*, topsep = 0 pt]
\item {\bfseries BERT4Rec}\cite{DBLP:conf/cikm/SunLWPLOJ19}: BERT4Rec is a session-based method adapted from the language model BERT~\cite{DBLP:conf/naacl/DevlinCLT19}. It employs bidirection information to model interactions which are not in a rigid order.
\item {\bfseries R-CE}\cite{DBLP:conf/wsdm/WangF0NC21}: An adaptive denoising training strategy (ADT) is applied for BERT4Rec to reduce the effect of hard interactions.
\item {\bfseries S$^3$-Rec}\cite{DBLP:conf/cikm/ZhouWZZWZWW20}: It utilizes mutual information maximization to capture intrinsic data correlation for sequential recommendation. For fair comparison, the extra attributes of items are removed, and the MIP and SP objectives are employed.
\item {\bfseries HyperRec}\cite{DBLP:conf/sigir/WangDH0C20}: It adopts hypergraph to model dynamic interactions between users and items in recommendation.
\end{itemize}
\noindent{\bfseries Event Prediction Methods:}
\begin{itemize}[leftmargin=*, topsep = 0 pt]
\item {\bfseries DuroNet-s}\cite{hu2021duronet}: It is a robust crime count prediction model that reduces the point-wise and the sequence-wise effect of noises. To adapt to our task, the spatial module is removed.
\end{itemize}
\noindent{\bfseries General Sequential Prediction Methods:}
\begin{itemize}[leftmargin=*, topsep = 0 pt]
\item {\bfseries Convolutional Self-attention (CSa)}\cite{DBLP:conf/nips/LiJXZCWY19}: It reduces the sensitivity to anomalies in series by utilizing causal convolution.
\item {\bfseries Informer}\cite{informer2021}: An efficient transformer-based model to capture dependence in extreme long sequences.
\end{itemize}

For BERT4Rec\footnote{https://github.com/FeiSun/BERT4Rec}, S$^3$-Rec\footnote{https://github.com/RUCAIBox/CIKM2020-S3Rec}, HyperRec\footnote{https://github.com/wangjlgz/HyperRec}, DuroNet-s\footnote{https://github.com/hukx-issac/DuroNet-for-crime-prediction}, and Informer\footnote{https://github.com/zhouhaoyi/Informer2020}, we use the code released by the authors. For CSa, we reproduce it in Pytorch. To make some regression methods adapt to our tasks, we add a shared embedding layer before and after the original model and adopt cross-entropy to train them. We adjust the hidden dimension size from \{32,64,128\}. The other hyper-parameters are set as reported in the papers. Their results are reported under the optimal settings.

\subsection{Overall Performance Comparison (RQ1)}
The comparison results with all baselines are shown in Table \ref{compbase}.
\begin{table*}[t]
  \caption{Accuracy comparison with baselines on four datasets. The optimal results are denoted in bold while the suboptimal results are underlined. "$*$" indicates significant improvement.}
	\resizebox{\linewidth}{!}{
  \begin{tabular}{clcccccccc||cccr}
    \toprule
    \multirow{2}{*}{Dataset}&		 \multirow{2}{*}{Metric}&				  	\multicolumn{5}{c}{Recommendation}&		\multicolumn{3}{c}{Non-recommendation}&	    \multicolumn{2}{c}{HAIL(ours)}& 		Diff.&	\multirow{2}{*}{Improv.}\\
\cmidrule(r){3-7}\cmidrule(r){8-10}\cmidrule(r){11-12}\cmidrule(r){13-13}
& &	POP&		BERT4Rec&	  			R-CE&	 S$^3$-Rec&	HyperRec&		DuroNet-s&		CSa&		\multicolumn{1}{c}{Informer}&  $\mathcal{M}_1$&  $\mathcal{M}_2$&	$\mathcal{M}_1$-$\mathcal{M}_2$& \\
     \hline
 	\hline
	\multirow{6}{*}{ML-1m}	&		HR@1&		0.0407&		0.3695& 	\underline{0.3988}&	0.2897&		0.3180&			0.1321&			0.1778&		0.0265&		{\bfseries 0.4291$^*$}&		0.4252&			+0.0039&				7.60\%		\\
						 	&		HR@5&		0.1603&		\underline{0.6851}& 	0.6478&	0.6575&		0.6631&			0.3849&			0.4629&		0.1154&		0.7202&		{\bfseries 0.7214$^*$}&			-0.0012&				5.30\%		\\
						  	&		HR@10&		0.2775&		0.7823& 				0.7404&	\underline{0.7911}&		0.7738&			0.5412&			0.6108&		0.2023&		0.8098&		{\bfseries 0.8146$^*$}&			-0.0048&				2.97\%		\\
						  	&		NDCG@5&		0.1008&		\underline{0.5375}& 	0.5327&	0.4557&		0.5014&			0.2616&			0.3243&		0.0707&		{\bfseries 0.5862$^*$}&		0.5843&			+0.0019&				9.16\%		\\
						  	&		NDCG@10&	0.1383&		\underline{0.5690}& 	0.5627&	0.5266&		0.5375&			0.3121&			0.3723&		0.0986&		{\bfseries 0.6155$^*$}&		0.6134&			+0.0021&				8.17\%		\\
						  	&		MRR&		0.1233&		\underline{0.5108}& 	0.5179&	0.4535&		0.4731&			0.2615&			0.3154&		0.0922&		0.5622&		{\bfseries 0.5791$^*$}&			-0.0169&				13.37\%		\\
	 \hline
	\multirow{6}{*}{Toys}	&		HR@1&		0.0260&		\underline{0.1390}& 	0.1130&	0.0990&		0.1147&			0.0465&			0.0534&		0.0144&		{\bfseries 0.1783$^*$}&		0.1780&			+0.0003&				28.27\%		\\
						  	&		HR@5&		0.1046&		\underline{0.3379}& 	0.3189&	0.3023&		0.2875&			0.1608&			0.1754&		0.0682&		0.3751&		{\bfseries 0.3755$^*$}&			-0.0004&				11.13\%		\\
						  	&		HR@10&		0.1848&		\underline{0.4596}& 	0.4529&	0.4393&		0.3909&			0.2572&			0.2723&		0.1286&		{\bfseries 0.4796$^*$}&		{\bfseries 0.4796$^*$}&	+0.0000&				4.35\%		\\
						  	&		NDCG@5&		0.0652&		\underline{0.2409}& 	0.2179&	0.2021&		0.2031&			0.1040&			0.1148&		0.0407&		{\bfseries 0.2802$^*$}&		{\bfseries 0.2802$^*$}&			+0.0000&				16.31\%		\\
						  	&		NDCG@10&	0.0909&		\underline{0.2802}& 	0.2611&	0.2463&		0.2365&			0.1350&			0.1459&		0.0600&		{\bfseries 0.3138$^*$}&		0.3136&			+0.0002&				11.99\%		\\
						  	&		MRR&		0.0861&		\underline{0.2444}& 	0.2233&	0.2081&		0.2087&			0.1211&			0.1301&		0.0628&		{\bfseries 0.2812$^*$}&		0.2810& 		+0.0002&			15.06\%		\\
	 \hline
	\multirow{6}{*}{CHI-18}	&		HR@1&		0.0249&		\underline{0.4421}& 	0.4114&	0.3978&		0.1679&			0.1386&			0.1378&		0.3507&		{\bfseries 0.4744$^*$}&		0.4703&			+0.0041&				7.31\%		\\
						  	&		HR@5&		0.1668&		\underline{0.6861}& 	0.6349&	0.6664&		0.3956&			0.4577&			0.4499&		0.5914&		{\bfseries 0.7117$^*$}&		0.7099&			+0.0018&				3.73\%		\\
						  	&		HR@10&		0.3250&		\underline{0.8024}& 	0.7708&	0.7942&		0.6088&			0.6356&			0.6333&		0.7166&		{\bfseries 0.8243$^*$}&		0.8228&			+0.0015&				2.73\%		\\
						  	&		NDCG@5&		0.0926&		\underline{0.5691}& 	0.5253&	0.5383&		0.2834&			0.3039&			0.2981&		0.4744&		{\bfseries 0.5985$^*$}&		0.5956&			+0.0029&				5.17\%		\\
						  	&		NDCG@10&	0.1440&		\underline{0.6068}& 	0.5692&	0.5799&		0.3525&			0.3613&			0.3569&		0.5148&		{\bfseries 0.6347$^*$}&		0.6321&			+0.0026&				4.60\%		\\
						  	&		MRR&		0.1190&		\underline{0.5567}& 	0.5197&	0.5243&		0.2957&			0.2961&			0.2917&		0.4650&		{\bfseries 0.5853$^*$}&		0.5823&			+0.0030&				5.14\%		\\
	 \hline
	\multirow{6}{*}{NYC-16}	&		HR@1&		0.0660&		0.4339& 	\underline{0.4472}&	0.3874&		0.3137&			0.1871&			0.1975&		0.3685&		{\bfseries 0.4772$^*$}&		0.4754&			+0.0018&				6.71\%		\\
						  	&		HR@5&		0.1994&		\underline{0.6931}& 	0.6261&	0.6909&		0.6358&			0.5401&			0.5509&		0.6178&		0.7166&		{\bfseries 0.7182$^*$}&			-0.0016&				3.62\%		\\
						  	&		HR@10&		0.3537&		0.8250& 				0.7088&	\underline{0.8287}&		0.7690&			0.7114&			0.7275&		0.7461&		 0.8433&		{\bfseries 0.8458$^*$}&			-0.0025&				2.06\%		\\
						  	&		NDCG@5&		0.1332&		\underline{0.5668}& 	0.5396&	0.5446&		0.4816&			0.3688&			0.3795&		0.4968&		0.6014&		{\bfseries 0.6019$^*$}&			-0.0005&				6.19\%		\\
						  	&		NDCG@10&	0.1824&		\underline{0.6095}& 	0.5665&	0.5893&		0.5247&			0.4244&			0.4363&		0.5383&		0.6427&		{\bfseries 0.6435$^*$}&			-0.0008&				5.58\%		\\
						  	&		MRR&		0.1616&		\underline{0.5527}& 	0.5322&	0.5245&		0.4611&			0.3511&			0.3608&		0.4862&		{\bfseries 0.5893$^*$}&		{\bfseries 0.5893$^*$}&			+0.0000&				6.62\%		\\
    \bottomrule
  \end{tabular}
}
\label{compbase}
\end{table*}
On the right side, we count the differences of accuracy between the base models $\mathcal{M}_1$ and $\mathcal{M}_2$, and the improvements of the best results relative to the suboptimal results. Several observations are summarized as follows:

\begin{figure*}[t]
  \centering
  \includegraphics[width=\linewidth]{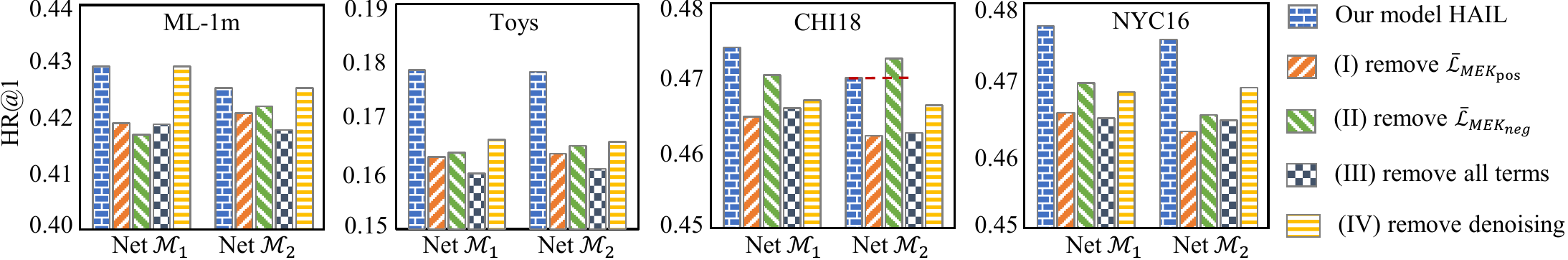}
	  \caption{Ablation Study of HAIL in terms of HR$@$1. The higher histograms, the better the performance.}
  \label {ablation}
  \vspace{-1 em}
\end{figure*}

For non-recommendation methods, the recommendation methods outperform them in most metrics. It is possibly caused by two reasons: $1)$ the non-recommendation methods generally assume a rigidly ordered sequence and design a relatively coarse-grain regression task to predict the next counts; $2)$ most recommendation methods conduct a cloze task \cite{DBLP:conf/naacl/DevlinCLT19} to pretrain their models, which generates more samples to train the models. Both DuroNet-s and CSa employ a convolution operator to reduce the effect of noises, which smooths the representations of related elements and reduces the differences. Hence, their values of HR$@1$ are significantly lower than other metrics. Informer is a specific method for extreme long sequences. However, it seems not suitable for recommendation scenario since it yields an inconsistent performance than POP in the recommendation datasets which lengths are short.

Among recommendation methods, BERT4Rec achieves comparable performance with S$^3$-Rec and HyperRec. It indicates that BERT4Rec makes use of interaction information without modeling generator. However, they do not outperform HAIL since they learn implicitly hard interactions under self-knowledge. When further changing the learning strategy of BERT4Rec with R-CE, the performance does not present significant improvement. It demonstrates that enhancing the learning of implicitly hard interaction is effective to enrich model patterns.

Finally, by comparing all the baselines, we can find that HAIL consistently achieves significant improvements. In terms of difference, the performance of model $\mathcal{M}_1$ is highly closed to that of model $\mathcal{M}_2$, indicating that exchanging mutual exclusivity knowledge can effectively reduce the gap between two models and improve generalization performance. For the dominant performance on datasets from different domains, HAIL is proved to be generalized to a much broader cyber or physical scenarios.

\subsection{Ablation Study (RQ2)}
To investigate the effectiveness of components in mutual exclusivity knowledge distillation, we remove the positive sample term~(variant~I), the negative sample term (variant II), both of them (variant III) in the Equation (\ref{total}) and the denoising strategy (variant~IV) in the Equation (\ref{denoise}), respectively. Note that the shared embedding and prediction layers are still kept. And then, we compare their performance with the original HAIL in terms of HR$@$1, since it is a harsh metric for performance evaluation. The results are reported in Figure \ref{ablation}. We have the following findings:
\begin{itemize}[leftmargin=*]
\item \emph{Finding 1: Mutual exclusivity knowledge distillation effectively improves the performance.} Compared with variant III, the performance of original model improves relatively 2.48\%, 11.51\%, 1.76\% and 2.67\% on four datasets, respectively. It demonstrates that enhancing the learning of implicitly hard interaction can improve the generalization performance of models.
\item \emph{Finding 2: The contribution of positive sample term is robust.} When removing the positive sample term, the performance of variant I
drops obviously. For example, compared with the original model, the performance of variant I decreases relatively 2.43\%, 9.52\%, 2.47\% and 2.51\% on four datasets. Meanwhile, the negative sample term shows a little fluctuation, particularly for CHI18 dataset (the red dash line). The observation demonstrates that distilling mutual exclusivity knowledge in positive samples is more efficient. The negative sample term might be affected by the size of elements, which denotes less candidates in prediction.
\item \emph{Finding 3: The denoising strategy proves to be helpful.} Except for the benchmark ML-1m dataset, denoising is imposed on the other datasets at different degrees. Compared with the original model, using denoising strategy improves the performance by 7.16\%, 1.50\% and 1.87\% on the three datasets. This is because denoising can effectively avoid the interference of noisy interactions.
\end{itemize}
\subsection{Parameter Sensitivity (RQ3)}
\begin{figure}[t]
  \centering
  \includegraphics[width=0.9\linewidth]{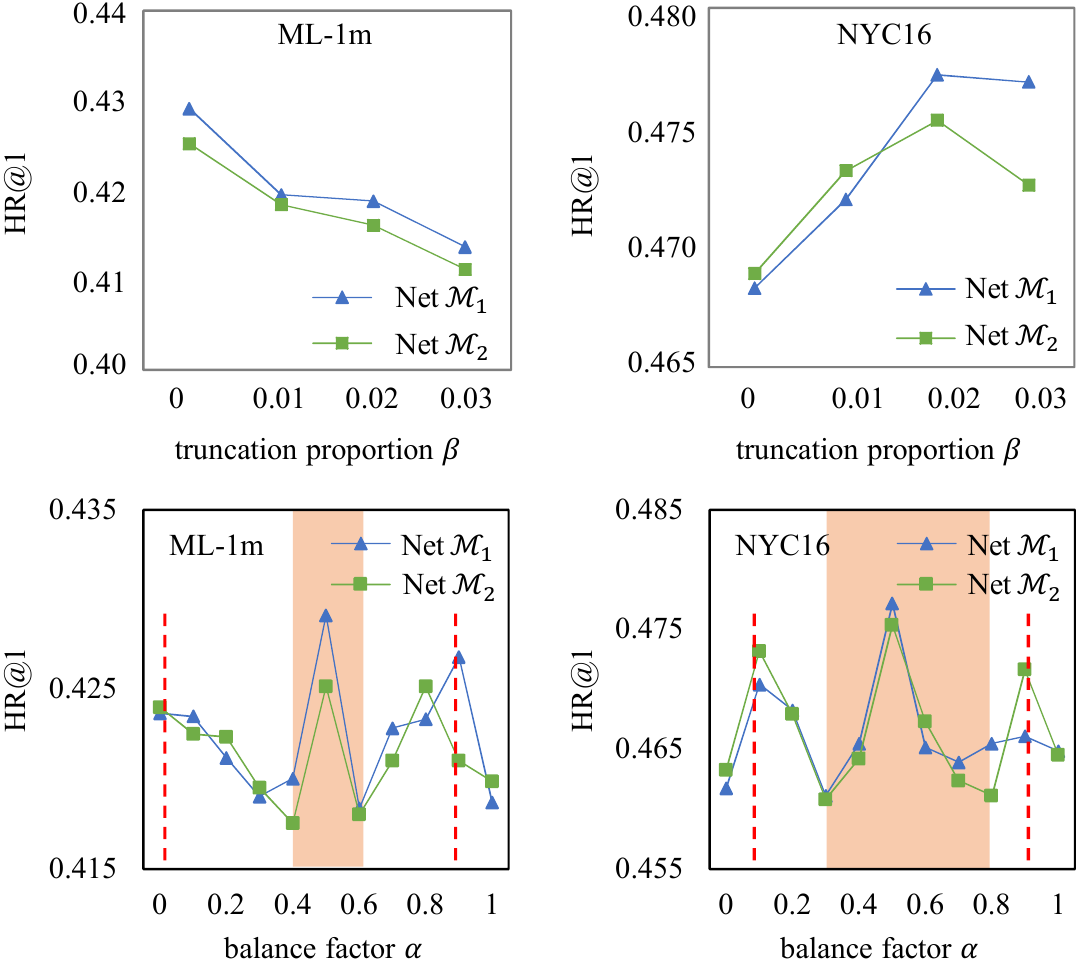}
	  \caption{The analysis of parameter sensitivity (HR$@$1).}
  \label {alpha}
\vspace{-1.3 em}
\end{figure}

To investigate the effect of different parameters, we tune the value of truncation proportion $\beta$ from 0 to 0.03 with a step 0.01 and the value of balance factor $\alpha$ from 0 to 1 with a step 0.1.

As shown in the first row of Figure \ref{alpha}, the bigger truncation proportion $\beta$, the greater strength of denoising. It can be observed that the performance of model directly falls or first rises and then falls with the growth of the value $\beta$. This is because the fitting of noisy interactions might mislead models. And, the performance can be improved if noises are removed. However, if limiting the noises too much, the meaningful hard interactions might be damaged and the performance decreases.

As shown in the second row of Figure \ref{alpha}, the bigger balance factor~$\alpha$, the smaller proportion of failure experience based loss. The red dash line indicates a meaningful region with a shape like "W". It implies that HAIL performs better when the ratio of two types of knowledge based learning is balanced or either of them achieves a dominated ratio. In fact, with the increase of failure experience based on loss ($\alpha$ varies from 1 to 0), the learning will step into three independent stages. To further explain this observation, the curve can be divided into three regions.
\begin{itemize}[leftmargin=*]
\item For the first stage in the right region, the proportion of self-knowledge based loss is dominated. Here, mutual exclusivity knowledge is similar to regularization, since some extra information about parameter learning can be introduced to lightly adjust the risk of overfitting.

\item For the center red region, both losses get into a balance period. In this stage, the base models can obtain a decent baseline from the self-knowledge and enhance the learning of implicitly hard interactions by employing mutual exclusivity knowledge. However, the performance decreases after the balance is broken.

\item As shown in the left region, the models mainly learn from mutual exclusivity knowledge. This stage is similar with the first stage where self-knowledge based loss is like a regularization. However, if its weight is set to 0, both the models might mislead each other.
\end{itemize}

\subsection{Evaluation w.r.t. Mutual Distillation}
To investigate the difference between mutual mimic learning \cite{DBLP:conf/cvpr/ZhangXHL18} and MED, two types of knowledge (i.e., likelihood distribution) mentioned in \S\ref{HIL} are transferred between two base models, respectively. The results from either of base models are randomly selected as the final output. As shown in Table \ref{kd}, the proposed MED is better than mutual mimic learning. It is because mutual mimic learning makes models prone to the learning of easy interactions while hard interactions might be taken for noises. However, sequential prediction is a challenging task with complex interaction patterns and substantial candidates. The proposed MED can effectively enhance the learning of implicitly hard interactions, which is more suitable for our task.

It is fair to discuss the increase of parameters and runtime. As shown in the left part of Figure \ref{framework}, this work tries to reduce parameters by introducing more shared layers. Parameters are only doubled in the base networks. In terms of the benchmark ML-1m dataset, the training time of mutual distillation is about 17 seconds for each batch while the individual model is about 9 seconds.

\begin{table}[tbp]
  \caption{The comparison between conventional knowledge distillation~(mimic learning) and the proposed MED.}
	\centering
\resizebox{0.9\linewidth}{!}{
  \begin{tabular}{ccccccc}
    \toprule
    \multirow{2}{*}{Dataset}& \multicolumn{3}{c}{Mutual mimic learning \cite{DBLP:conf/cvpr/ZhangXHL18}}&		\multicolumn{3}{c}{MED}\\
	\cmidrule(r){2-4}\cmidrule(r){5-7}
					& HR@1&			NDCG@5&		MRR&		HR@1&			NDCG@5&		MRR			\\
	\midrule
    ML-1m& 			0.3952&			0.5656&		0.5386&		0.4291&			0.5862&		0.5622		\\
	Toys&			0.1693&			0.2761&		0.2767&		0.1783&			0.2802&		0.2812		\\
	CHI-18&			0.4699&			0.5932&		0.5815&		0.4744&			0.5985&		0.5853		\\
	NYC-16&			0.4660&			0.5956&		0.5806&		0.4772&			0.6014&		0.5893		\\
  \bottomrule
\end{tabular}
\label {kd}
}
\vspace{-1.5 em}
\end{table}

\section{Conclusion and Future Work}
In this paper, we highlight the effect of implicitly hard interactions. To this end, a hardness aware interaction learning framework called HAIL is proposed to enhance the learning of them. In particular, based on the proposed MED, implicitly hard interactions are identified from different perspectives. And, both base models derive training experience from each other to adjust themselves learning strategy. Extensive experiments are conducted on four datasets covering cyber and physical spaces. The results show that HAIL outperforms several state-of -the-art methods. For future work, we are interested in extending the framework to more complex sequential prediction tasks, such as multi-modal prediction, and sequence-to-sequence prediction.

\begin{acks}
This work is partially supported by the China Scholarship Council (LiuJinMei [2020] 1509, 202106950041).
\end{acks}

\bibliographystyle{ACM-Reference-Format}
\balance
\bibliography{HAIL}


\end{document}